\PassOptionsToPackage{usenames,table,dvipsnames}{xcolor}
\documentclass{article}
\usepackage[final]{nips_2017}
\usepackage[utf8]{inputenc} 
\usepackage[T1]{fontenc}    
\usepackage{hyperref}       
\usepackage{url}            
\usepackage{booktabs}       
\usepackage{amsfonts}       
\usepackage{nicefrac}       
\usepackage{microtype}      
\usepackage{graphicx}
\usepackage{amsmath}
\usepackage{tikz}
\usetikzlibrary{positioning, shapes.geometric, arrows.meta}
\usepackage{natbib}
\usepackage{hyperref}
\usepackage{float}
\usepackage{xcolor}
\usepackage{tcolorbox}
\usepackage{xcolor,soul}
\sethlcolor{cyan!30} 

\title{Automated LaTeX Code Generation from Handwritten Mathematical Expressions}

\author{
  Jayaprakash Sundararaj \\
  \texttt{osjp@stanford.edu} \footnote{This research conducted as part of Stanford AI Curriculum.} \\
  \AND
  Akhil Vyas \\
  \texttt{avyas21@stanford.edu}  \\
  \AND
  Benjamin Gonzalez-Maldonado \\
  \texttt{bengm@stanford.edu } \\
}

\begin{document}

\maketitle

\begin{abstract}
Transforming mathematical expressions into LaTeX poses a significant challenge. In this paper, we examine the application of advanced transformer-based architectures to address the task of converting handwritten or digital mathematical expression images into corresponding LaTeX code. As a baseline, we utilize the current state-of-the-art CNN encoder and LSTM decoder. Additionally, we explore enhancements to the CNN-RNN architecture by replacing the CNN encoder with the pretrained ResNet50 model with modification to suite the grey scale input. Further, we experiment with vision transformer model and compare with Baseline and CNN-LSTM model. Our findings reveal that the vision transformer architectures outperform the baseline CNN-RNN framework, delivering higher overall accuracy and BLEU scores while achieving lower Levenshtein distances. Moreover, these results highlight the potential for further improvement through fine-tuning of model parameters. To encourage open research, we also provide the model implementation, enabling reproduction of our results and facilitating further research in this domain \footnote{\href{https://github.com/osjayaprakash/deeplearning/tree/main}{https://github.com/osjayaprakash/deeplearning/tree/main}}.
\end{abstract}

\section{INTRODUCTION}

Converting handwritten mathematical expressions into digital formats, particularly LaTeX code, is a complex and time-consuming task. LaTeX is widely used for typesetting mathematical documents, but manually transcribing handwritten notes into LaTeX is tedious and prone to errors. Our objective is to develop a machine learning (ML) model that can seamlessly transform handwritten mathematical expressions into accurate LaTeX source code, streamlining this process significantly.

The input to our proposed algorithm is an image containing a handwritten mathematical expression. This image is processed by the model to produce a sequence of tokens that represent the equivalent LaTeX code. Achieving this transformation requires addressing significant challenges, as the task involves elements of both computer vision and natural language processing (NLP). Computer vision is essential for interpreting and understanding the visual structure of the handwritten expression, while NLP techniques are crucial for generating the textual LaTeX sequence from the extracted features.

To address these challenges, we employ an encoder-decoder architecture. The encoder processes the input image, extracting meaningful features that capture the underlying mathematical structure. The decoder then converts these features into a sequential representation of the LaTeX code. This architecture allows us to bridge the gap between visual input and textual output, paving the way for efficient mathematical transcription.

\section{RELATED WORK}

Recent advancements in converting handwritten or image-based mathematical expressions into digital formats have been driven by a variety of methods, each addressing the inherent complexities of the task. In earlier works, Schechter et al. \cite{schechter2017converting} explored multiple approaches, including neural networks, convolutional neural networks (CNNs), Random Forests, Support Vector Machines (SVMs), Optical Character Recognition (OCR) systems, Conditional Random Fields (CRF), and Sequence Alignment (SA). While these approaches laid the groundwork, they lacked the sophistication required for handling the intricate structure of mathematical expressions, particularly the spatial dependencies between symbols.

State-of-the-art methods have largely converged on encoder-decoder architectures that incorporate CNNs and Recurrent Neural Networks (RNNs). Genthial et al. \cite{genthial2016image} demonstrated the effectiveness of these architectures in capturing spatial features using CNN encoders and generating sequential outputs with Long Short-Term Memory (LSTM) decoders. This framework has since become a standard baseline for many image-to-sequence tasks. More recent advancements, such as the work by Bian et al. \cite{bian2022handwritten}, extended the encoder-decoder paradigm by incorporating both left-to-right and right-to-left decoders, enhancing the ability to handle complex structures and dependencies in mathematical expressions.

Despite the success of CNN-RNN architectures, challenges remain, particularly in capturing long-range dependencies and representing input features efficiently. Transformer architectures \cite{transformer}, which excel in Natural Language Processing (NLP) tasks, offer a promising alternative. These architectures rely on self-attention mechanisms, enabling them to model relationships between distant elements in a sequence effectively. By eschewing recurrence, transformers mitigate the limitations of RNNs, such as vanishing gradients and sequential dependency constraints.

The introduction of Vision Transformers (ViTs) \cite{visiontransformer} marked a significant milestone in adapting transformers for computer vision tasks. Unlike traditional CNNs, which rely on convolutional layers to extract local features, vision transformers divide input images into a sequence of patches and process them using self-attention mechanisms. This approach allows the model to capture both local and global features simultaneously, resulting in superior performance on image-based tasks. Furthermore, the inherent flexibility of transformers enables their adaptation to various input modalities, making them particularly well-suited for tasks that require multimodal processing.

Given the promising results of transformers in NLP and computer vision, our work leverages a vision transformer encoder combined with a transformer decoder to address the problem of converting handwritten mathematical expressions into LaTeX. This architecture builds on the strengths of vision transformers for feature extraction and the transformer decoder's ability to generate high-quality sequential outputs. By comparing this approach with the CNN-RNN baseline, we aim to demonstrate the efficacy of transformer-based models for this task.

In summary, while traditional methods like CNN-RNN architectures have provided a solid foundation for mathematical expression recognition, the emergence of transformer architectures represents a paradigm shift. By incorporating self-attention mechanisms and handling sequential and spatial data more effectively, transformers offer the potential to achieve state-of-the-art performance, paving the way for further advancements in this domain.

\section{DATASET AND FEATURES}

We will use the datasets from two main repositories: \texttt{Im2latex-100k} (\cite{kanervisto_2016_56198}) and \texttt{Im2latex-230k} (\cite{gervais2024mathwritingdatasethandwrittenmathematical}). The \texttt{Im2latex-100k} (\cite{kanervisto_2016_56198}) dataset, available at \href{https://zenodo.org/records/11230382}{Zenodo}, contains 100,000 image-formula pairs. The \texttt{Im2latex-230k} (\cite{gervais2024mathwritingdatasethandwrittenmathematical}) 
 dataset, also known as \texttt{Im2latexv2}, contains 230,000 samples. It includes both OpenAI-generated and handwritten examples, further enhancing the diversity of the data. This dataset is available at \href{https://im2markup.yuntiandeng.com/data/}{Im2markup}. The training data format is \texttt{<image file name> <formula id>}.
 
\begin{figure}[H]
    \centering
    \includegraphics[scale=0.4]{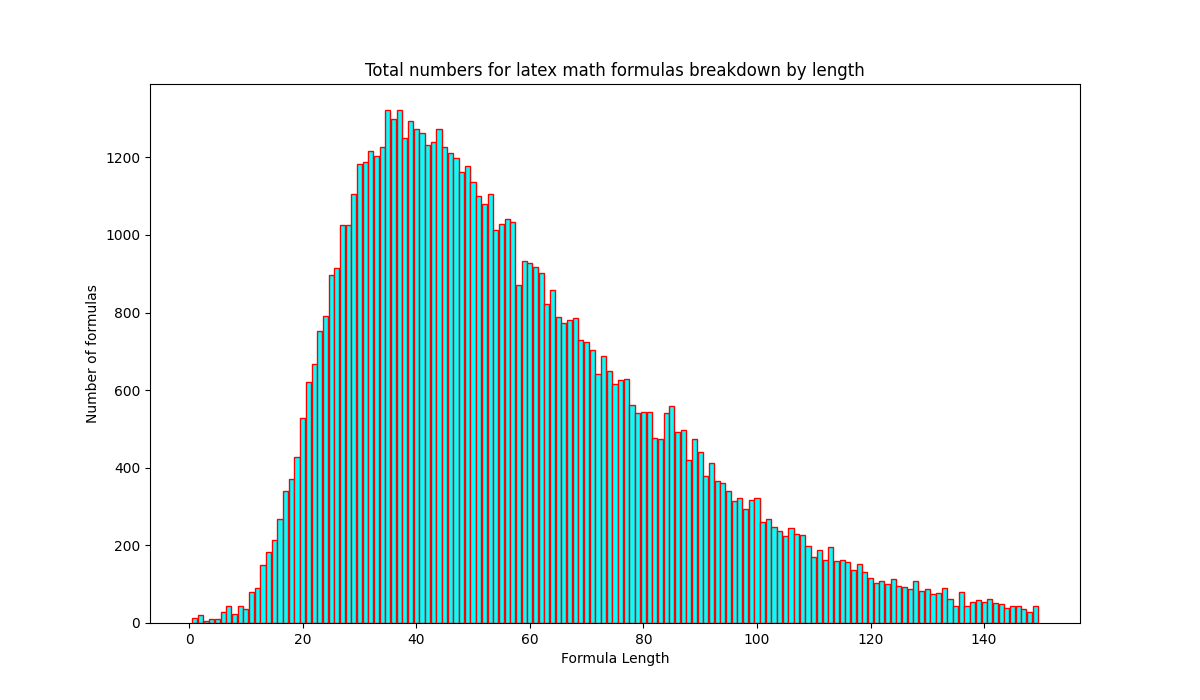}
    \caption{Formulas breakdown by length}
    \label{fig:formula_length}
\end{figure}

\begin{figure}[H]
    \centering
    \includegraphics[scale=0.4]{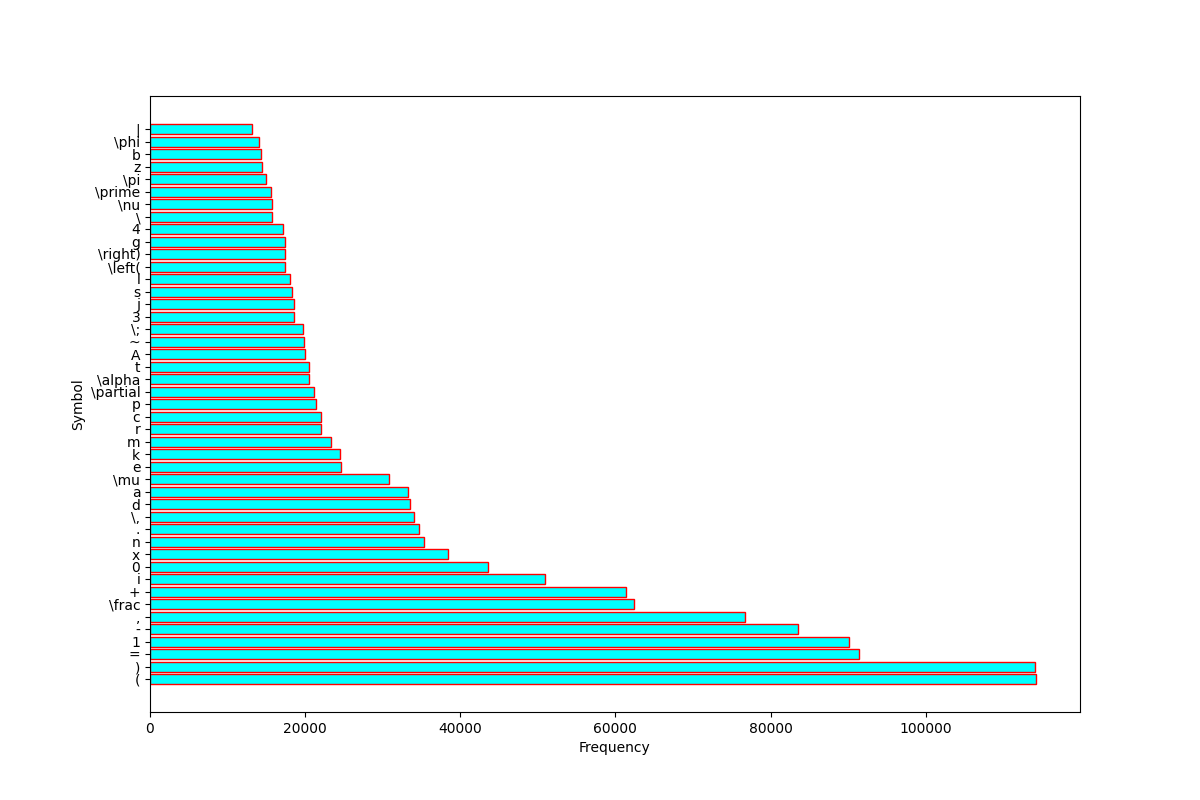}
    \caption{Dataset: Most popular symbols and frequencies.}
    \label{fig:vocab_freq_1}
\end{figure}

\begin{figure}[H]
    \centering
    \includegraphics[scale=0.4]{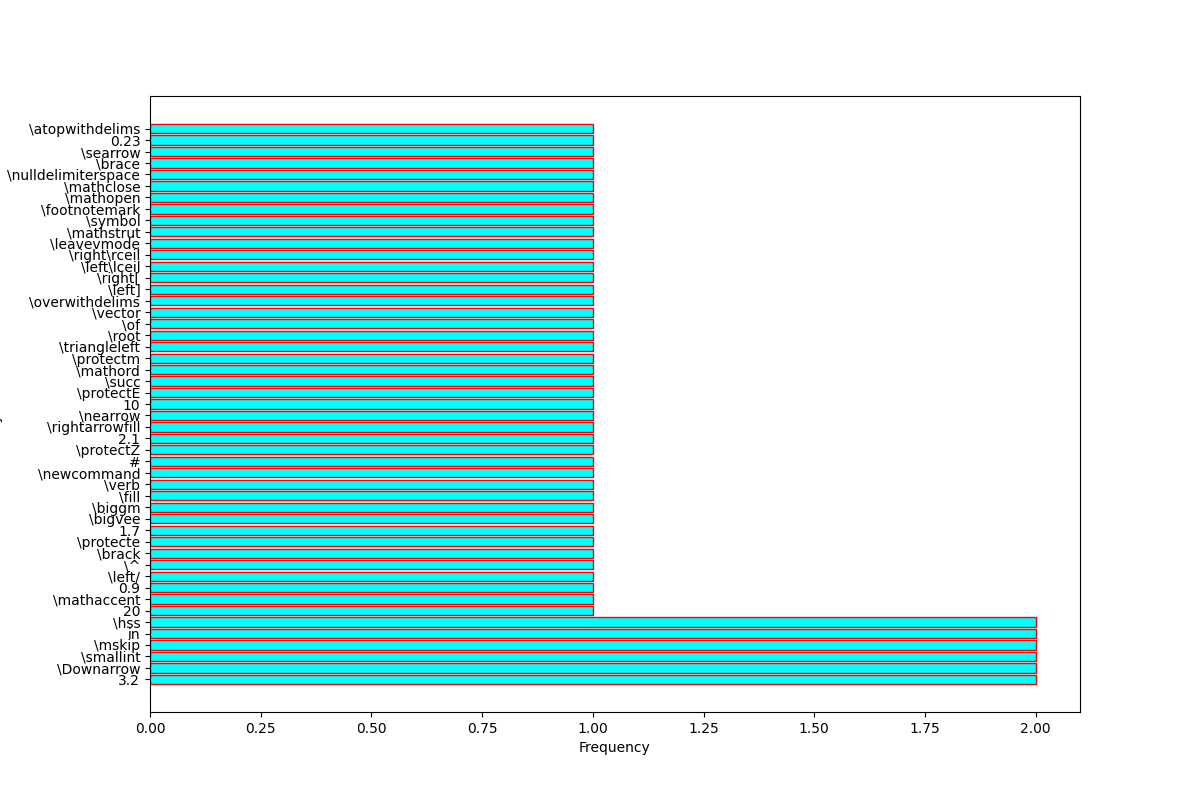}
    \caption{Dataset: Least popular symbols and frequencies.}
    \label{fig:vocab_freq_2}
\end{figure}

 The dataset disk size is 849 MB. The images are gray scales with 50x200 pixels. The numbers of symbols (\autoref{fig:formula_length}) in the latex formulas vary from range varies from 1 to 150 symbols. Voabulary contains 540 symbols, refer \autoref{fig:vocab_freq_1} and \autoref{fig:vocab_freq_2} for the list of popular and least occurring symbols with their frequency.
 
\section{METHODS}

\subsection{BASELINE: CNN ENCODER AND LSTM DECODER}
As a baseline, we employ a CNN encoder to process the input image, which is resized to dimensions of 50×200 pixels with a single channel (grayscale). The architecture utilizes a 3×3 convolutional filter, followed by a 2×2 max-pooling layer. This block is repeated three times to progressively extract features from the image. The final output from the convolutional layers is passed through a fully connected layer to complete the encoding process.

\begin{figure}[H]
    \centering
    \includegraphics[scale=0.7]{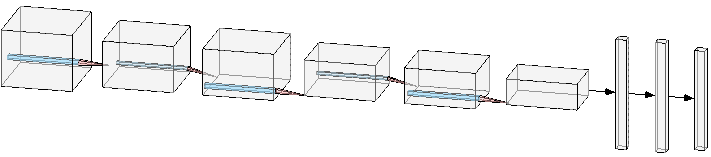}
    \caption{Encoder architecture consists of 3 convolution-max pooling blocks (50,200) -> (25,100) -> (12,5) which is flattened and fed into Dense layer (256 units)  }	
    \label{fig:cnn_lstm}
\end{figure}

During the decoding phase, embeddings for the formula tokens are computed and concatenated with the image-encoded embeddings. This combined embedding, which integrates both image and token information, is then passed into LSTM or GRU units. The output from these recurrent units is processed through a fully connected network, with a softmax activation function applied to generate the final predictions.

The overall model architecture can be summarized as follows:

\begin{tikzpicture}[
    node distance=1cm, 
    every node/.style={rectangle, draw, minimum height=1.2cm, minimum width=1cm, align=center},
    arrow/.style={-Stealth, thick}
]

\node (input) [rounded corners] {Input Image \(I\)};
\node (cnn) [right=0.6cm of input] {CNN (Encoder)};
\node (lstm) [right=0.6cm of cnn] {LSTM/GRU (Decoder)};
\node (output) [right=0.6cm of lstm, rounded corners] {Output LaTeX Sequence \(T\)};

\draw [arrow] (input) -- (cnn);
\draw [arrow] (cnn) -- (lstm);
\draw [arrow] (lstm) -- (output);

\end{tikzpicture}

\subsection{FINETUNING PRETRAINED RESENT50 MODEL}

In this experiment, we utilize the pretrained ResNet50 model as the encoder, which has a disk size of 98 MB. ResNet50 is a well-established architecture known for its robust performance on various computer vision tasks, owing to its pretraining on large-scale datasets. However, the ResNet50 model expects input images of a fixed size, specifically $254 \times 254$ pixels, with three color channels (RGB). This poses a challenge since our input images are in grayscale format, with a single channel.

\begin{figure}[H]
    \centering
    \includegraphics[scale=0.45]{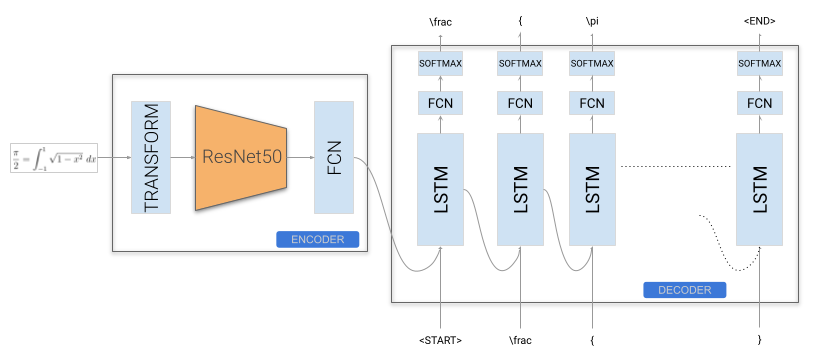}
    \caption{Pretrained ResNet50 Encoder with LSTM Decoder.}
    \label{fig:resnet_lstm}
\end{figure}

To address this mismatch, we transform the grayscale input images into the required RGB format. This is achieved using the \texttt{tf.image.grayscale\_to\_rgb} function provided by TensorFlow. Specifically, we apply this transformation through a Lambda layer in Keras, as shown below:

\begin{center}
\begin{verbatim}
tf.keras.layers.Lambda(lambda x: tf.image.grayscale_to_rgb(x))
\end{verbatim}
\end{center}

This transformation replicates the grayscale channel across the three RGB channels, allowing the ResNet50 model to process the input images as expected. Additionally, the images are resized to match the fixed input dimensions of ResNet50, ensuring compatibility with the architecture. By leveraging this approach, we are able to utilize the pretrained ResNet50 encoder effectively, benefiting from its rich feature extraction capabilities while maintaining the integrity of our grayscale input data.

\subsection{VISION TRANSFORMER}

\subsubsection{ENCODER}

The vision encoder in our architecture processes the input image by dividing it into smaller patches, a technique inspired by the Vision Transformer (ViT) framework. This patch-based approach enables the model to capture local spatial information efficiently while leveraging self-attention mechanisms to analyze global dependencies across the entire image.

For our implementation, we create patches of size \(10 \times 10\) pixels. Given that the input images have dimensions of \(50 \times 200\) pixels, the image is divided into a total of 100 patches. The total number of patches is computed as:

\[
\text{Total patches} = \frac{\text{Height}}{\text{Patch height}} \times \frac{\text{Width}}{\text{Patch width}} = \frac{50}{10} \times \frac{200}{10} = 5 \times 20 = 100
\]

Each patch is treated as a flattened vector, which is then passed through a learnable embedding layer. This embedding layer maps the patch representation into a high-dimensional space suitable for processing by the transformer encoder. By representing the image as a sequence of patches rather than using traditional convolutional layers, the model effectively captures both local and global features, improving its ability to handle complex visual structures.

This patch-based encoding approach is particularly advantageous for tasks requiring fine-grained analysis and flexibility in handling varying input dimensions, making it a crucial component of the vision encoder in our system.

\begin{figure}[h]
	\centering \includegraphics[scale=0.6]{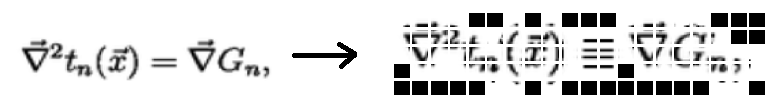}
	\caption{Original latex image and the generated patches}
\end{figure}

In the Vision Transformer (ViT) encoder, the image patches are first processed through a linear embedding layer, transforming each patch into a fixed-dimensional vector. These embeddings are then augmented with positional embeddings to retain spatial information about the arrangement of patches within the image. This combined representation is subsequently fed into a standard transformer layer for further processing.

Our architecture employs a total of 8 transformer layers. Each layer consists of 4 attention heads, enabling the model to focus on different parts of the input simultaneously. Additionally, each transformer layer includes a multi-layer perceptron (MLP) block with two layers, consisting of 2048 and 1024 units, respectively. This MLP block applies non-linear transformations to enhance the model's representational capacity. The overall structure is designed to capture both local and global dependencies effectively, making it well-suited for complex image-to-sequence tasks.

\begin{figure}[H]
	\centering \includegraphics[scale=0.6]{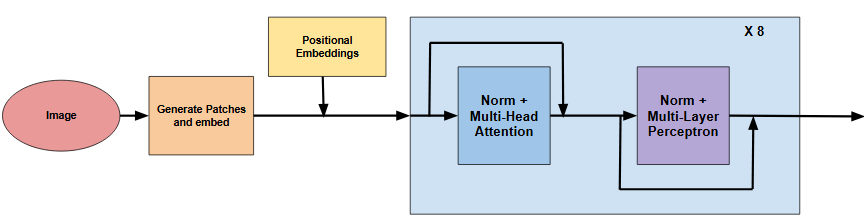}
	\caption{Transformer encoder architecture}
\end{figure}

\subsubsection{DECODER}

For the decoder, we use a standard transformer block that incorporates both cross-attention and self-attention mechanisms. The cross-attention component is utilized to identify and focus on specific regions of the image that are most relevant to the current stage of decoding. This enables the decoder to effectively align the input image features with the sequence being generated. Concurrently, the self-attention mechanism is used to model dependencies within the generated sequence, ensuring coherence and context-awareness across the output tokens.

Our configuration consists of 4 attention layers in the decoder. Each layer is equipped with 8 attention heads for both the cross-attention and self-attention components. This setup allows the decoder to process multiple aspects of the input and output simultaneously, enhancing its ability to generate accurate and well-structured sequences. This combination of cross-attention and self-attention in a multi-head configuration is crucial for effectively bridging the gap between the visual input and the sequential output in our architecture.


\section{EXPERIMENTS}

\subsection{SETUP AND HYPERPARAMETERS}

The models were trained on 200,000 data points using a single AWS G6.xlarge instance equipped with an Nvidia Tesla GPU. The training duration ranged from 1 hour 30 minutes to 2 hours, depending on the architecture. Early stopping with a patience of 10 epochs was implemented to prevent overfitting and ensure efficient training.

For the CNN-LSTM and ResNet-LSTM models, a batch size of 128 was used, while the transformer architecture employed a smaller batch size of 64 due to GPU memory constraints inherent to the AWS instance configuration. The choice of batch sizes was carefully calibrated to balance memory usage and training efficiency.

The learning rates and optimization strategies were tailored to each model. The CNN-LSTM and ResNet-LSTM models utilized the Adam optimizer with a default learning rate of 0.001, which provided stable and effective convergence. In contrast, the transformer architecture leveraged the AdamW optimizer, employing a learning rate schedule that decayed from 1e-4 to 1e-6. This learning rate decay strategy was determined through experimentation and was found to enhance the performance and stability of the transformer model during training. These configurations were designed to optimize the performance of each model within the constraints of the computational resources available.

\subsection{METRICS}

We compute the following metrics to compare the baseline and other methods:
\begin{itemize}
  \item We measure the `sparse categorical loss` and accuracy which measures the loss/accuracy accross all tokens. 
  \item We measure the masked loss and accuracy to measure the accuracy for the non-padded tokens (we pad our tokens to length 151 which is the max and this will only check the loss/accuracy for the tokens that are part of the label sequence)
  \item We measure the Levenshtein distance and BLEU-4 score for predicted sequences of a subset of the training set. These metrics were chosen in order to quantify closeness/correctness between sequences beyond a simple binary score that relies on exact matching.
\end{itemize}

\subsection{LSTM AND RNN BASELINE COMPARISION}
We explored using both LSTM/GRU for the decoder and the difference between the two were negligible. Here are the training curves with CNN - LSTM/GRU architectures:
\newline \newline
 \includegraphics[scale=0.45]{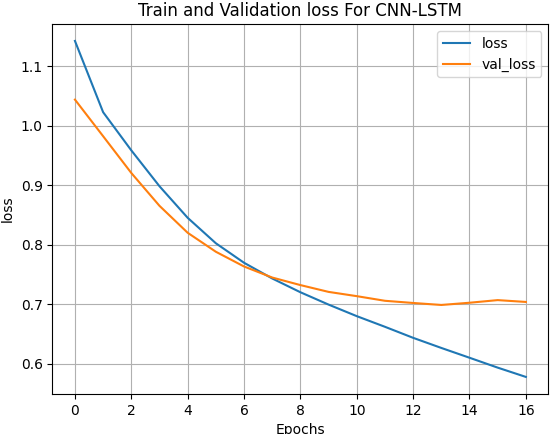}
 \includegraphics[scale=0.45]{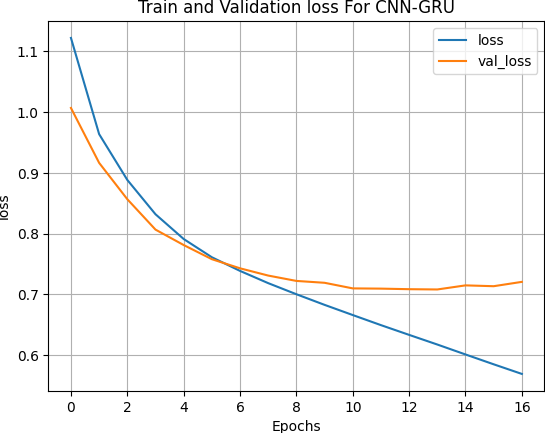}

Both models had ~85\% accuracy with GRU being slightly worse off. Due to this, we used the numbers from the LSTM decoder to compare against the other models. 

\subsection{RESULTS}

The experimental results indicate that the vision transformer architecture achieves significantly lower loss and Levenshtein scores, as well as higher accuracy and BLEU scores, when compared to the baseline CNN-LSTM model and the ResNet-LSTM model. This highlights the superior performance of vision transformers in capturing complex features and dependencies within the data.

\begin{center}
\begingroup
\setlength{\tabcolsep}{8pt} 
\renewcommand{\arraystretch}{2} 
\begin{tabular}{|c | c | c | p{15mm} | p{15mm} | c | c |} 
 \hline
 \rowcolor{NavyBlue!90}
 \textcolor{white}{Architecture} &  \textcolor{white}{Loss} &  \textcolor{white}{Accuracy} &  \textcolor{white}{Masked \newline  Loss} &  \textcolor{white}{Masked \newline Accuracy} &  \textcolor{white}{BLEU} &  \textcolor{white}{Levenshtein} \\ [0.5ex] 
 \hline
 CNN-LSTM & 0.647 & 0.847 & 1.694 & 0.600 & 0.429 &  0.401 \\ 
 \hline
ResNet-LSTM  & 0.781  & 0.819 & 2.079  & 0.520 & 0.372  &  0.435 \\ 
 \hline
Vis Transformer  & 0.520 & 0.873 & 1.472 & 0.641 & 0.557 & 0.354 \\
 \hline
\end{tabular}
\endgroup
\end{center}

The ResNet-LSTM model also demonstrated improved performance over the CNN-LSTM model. This can be attributed to the fact that ResNet benefits from pretraining on large corpora of vision-related datasets, which enables it to encode a rich understanding of visual semantics. The pretrained embeddings in ResNet facilitate faster convergence during training, as the model leverages the prelearned representations rather than learning features entirely from scratch.

In contrast, the CNN-LSTM and Vision Transformer models must learn both the feature extraction and sequence modeling tasks from the ground up. This lack of pretraining results in slower convergence and limits their ability to generalize effectively. These findings underscore the advantages of transfer learning and pretrained architectures, particularly in tasks involving complex image-to-sequence mappings, as they provide a strong initialization point and accelerate the training process.

Overall, the experiments demonstrate the transformative potential of vision transformer architectures and pretrained models in addressing challenges in mathematical expression recognition and similar tasks.

\section{CONCLUSION AND FUTURE WORK}
The results demonstrate that the transformer architecture consistently outperformed the vanilla CNN-RNN architecture across all evaluated metrics. We attribute this superior performance to the algorithm's effective use of positional embeddings combined with attention mechanisms in both the encoder and decoder components.

With additional time, our efforts would have been directed towards further exploration of various transformer architecture configurations, including adjustments to the number of layers, attention heads, and patch sizes in the vision transformer encoder. Additionally, we would have focused on training more complex models with larger datasets to further enhance performance.

For this study, we utilized a dataset comprising 200,000 samples, paired with appropriately sized models, to accommodate the GPU memory limitations of the AWS instances available to us. These constraints influenced the scope of our experimentation but nonetheless provided valuable insights into the potential of transformer-based architectures for this task.

\nocite{*}
\bibliography{paper}

%
%
%

\newpage

\section{APPENDIX}

\end{document}